\documentclass[letterpaper, 10 pt, conference]{ieeeconf}  % Comment this line out if you need a4paper

\IEEEoverridecommandlockouts                              % This command is only needed if 
\usepackage[utf8]{inputenc}
\usepackage{hyperref}
\usepackage{url}
\usepackage{amsfonts}
\usepackage{amssymb}
\usepackage{amsmath}
\usepackage{dsfont}
\usepackage{algorithm}
\usepackage{algorithmicx}
\usepackage{xcolor}
\usepackage{comment}
\usepackage{graphicx}
\usepackage{physics}
\usepackage{multirow}
\usepackage{tikz}
\usepackage{ifthen}
\usepackage{adjustbox}
\usepackage{lipsum}
\usepackage{tabularx}
\usepackage{calc}

\usepackage{svg}
\usepackage{cite}

\title{\LARGE \bf
Self-Improving Safety Performance of Reinforcement Learning Based Driving with Black-Box Verification Algorithms
}

\author{Resul Dagdanov$^{\ast}$, Halil Durmus, and Nazim Kemal Ure%
\thanks{$^\ast$ The corresponding author.}% <-this % stops a space
\thanks{R. Dagdanov is with ITU Artificial Intelligence and Data Science Research Center and the Department of Aeronautical Engineering,
        Istanbul Technical University, Turkey
        {\tt\small dagdanov21 at itu.edu.tr}}%}%
\thanks{H. Durmus is with Eatron Technologies and the Department of Electronics and Communication Engineering,
        Istanbul Technical University, Turkey
        {\tt\small durmush at itu.edu.tr}}%}%
\thanks{N.K. Ure is with ITU Artificial Intelligence and Data Science Application and Research Center and Department
        of Computer Engineering, Istanbul Technical University, Turkey
        {\tt\small ure at itu.edu.tr}}%
}

\begin{document}

\maketitle
\thispagestyle{empty}
\pagestyle{empty}

%%%%%%%%%%%%%%%%%%%%%%%%%%%%%%%%%%%%%%%%%%%%%%%%%%%%%%%%%%%%%%%%%%%%%%%%%%%%%%%%%%%%%%%%%%%%%%%%
% ------------------------------------------ ABSTRACT ------------------------------------------ 
%%%%%%%%%%%%%%%%%%%%%%%%%%%%%%%%%%%%%%%%%%%%%%%%%%%%%%%%%%%%%%%%%%%%%%%%%%%%%%%%%%%%%%%%%%%%%%%%
\begin{abstract}
In this work, we propose a self-improving artificial intelligence system to enhance the safety performance of reinforcement learning (RL)-based autonomous driving (AD) agents using black-box verification methods. RL algorithms have become popular in AD applications in recent years. However, the performance of existing RL algorithms heavily depends on the diversity of training scenarios. A lack of safety-critical scenarios during the training phase could result in poor generalization performance in real-world driving applications. We propose a novel framework in which the weaknesses of the training set are explored through black-box verification methods. After discovering AD failure scenarios, the RL agent's training is re-initiated via transfer learning to improve the performance of previously unsafe scenarios. Simulation results demonstrate that our approach efficiently discovers safety failures of action decisions in RL-based adaptive cruise control (ACC) applications and significantly reduces the number of vehicle collisions through iterative applications of our method. The source code is publicly available at \url{https://github.com/data-and-decision-lab/self-improving-RL}.
\end{abstract}

%%%%%%%%%%%%%%%%%%%%%%%%%%%%%%%%%%%%%%%%%%%%%%%%%%%%%%%%%%%%%%%%%%%%%%%%%%%%%%%%%%%%%%%%%%%%%%%%
% ---------------------------------------- INTRODUCTION ---------------------------------------- 
%%%%%%%%%%%%%%%%%%%%%%%%%%%%%%%%%%%%%%%%%%%%%%%%%%%%%%%%%%%%%%%%%%%%%%%%%%%%%%%%%%%%%%%%%%%%%%%%
\section{Introduction}
Fully reliable and comfortable autonomy applications increasingly prioritize safety-critical implementations in autonomous systems. The advancement of autonomous driving (AD) technologies is largely due to the widespread use of the black-box systems such as deep neural networks in numerous control \cite{training} and planning \cite{dynamic_occlusion} problems. To ensure both environmental and human safety, black-box systems must undergo comprehensive testing and validation in a variety of edge scenarios before they can be deployed in real-world use cases.

In a recent study \cite{safedagger}, researchers reduced unsafe scenarios of a black-box system by guiding exploration samples along predefined trajectory classes. However, without verification through rare-event simulation \cite{stochastic} and generalized importance sampling on a continuous action and observation space, the safety of the AD system could not be guaranteed. In another safety-critical system, researchers proposed a safety-constrained collision avoidance approach \cite{reachability} with prediction-based reachability analysis. Nevertheless, these approaches do not verify the designed models on the continuous feature domain of scenarios. As the operating environment of the black-box system becomes more complicated, testing the control policy in all potential circumstances becomes impractical. Hence, innovative techniques are needed to verify the safety performance of the black-box system. In the context of AD, stress testing methods for rare-event probability validations are used, such as adaptive multilevel splitting (AMS) \cite{efficient}, reinforcement learning (RL) \cite{adaptive_stress}, Bayesian optimization \cite{generating_adversarial}, cross-entropy \cite{scalable}, unsupervised clustering \cite{adaptive_generation}, neural bridge sampling \cite{neural_bridge}, and genetic algorithm \cite{interpretable}. Although these applications identify potential black-box system failures, they do not make use of the discovered occurrences to improve or repair the safety-critical system. In this work, our main focus is on leveraging rare-event sampling approaches to gain better insights into failure instances, which could be made feasible with a lesser number of iterations.

Consider a typical AD scenario where the EGO vehicle comfortably and safely pursues the lead vehicle or most important object (MIO). In ideal adaptive cruise control (ACC) settings, an EGO vehicle is expected to follow the MIO vehicle with a comfortable braking distance, the least amount of jerk, and pleasant acceleration. Without any prior knowledge of the mathematical model of the system, ACC is effectively approached using model-free deep RL through trial and error on the safety and comfort parameters in interaction with the simulation environment \cite{saint_acc}. Similarly, another study \cite{lane_change} used deep RL to safely enable lane-changing maneuvers and demonstrated the advantages of model-free algorithms compared to rule-based baselines \cite{idm}. However, the safety of these trained RL agents is not fully verified over the continuous range of rare events, where time-to-collision (TTC) measurements are low and the probability of collision is high. In another conducted study \cite{avoid_collisions}, the effects of TTC on the high likelihood of collisions were analyzed using driver participants in simulation environments.

\begin{figure*}[t!]
    \centering
    \includegraphics[width=0.95\textwidth]{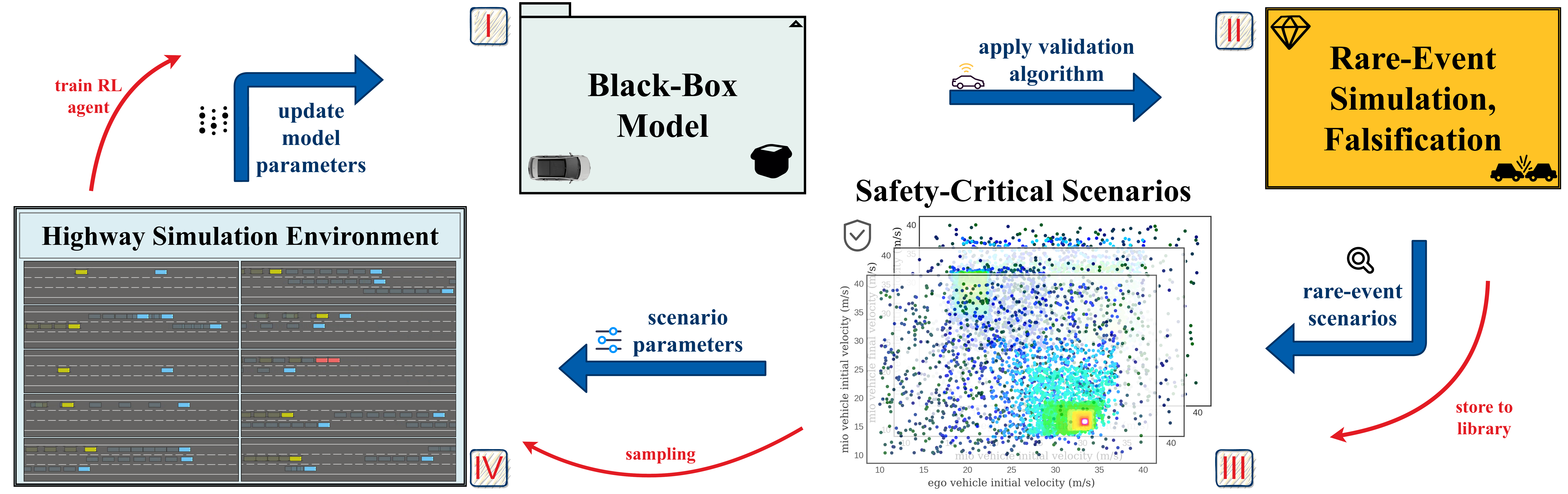}
    \caption{{\bfseries{Proposed Method Overview}}. {\bfseries{Stage-I}}: A black-box model is trained in a simulation environment for predefined time steps. {\bfseries{Stage-II}}: The trained RL agent is validated using a rare-event simulation algorithm. {\bfseries{Stage-III}}: Estimated scenario parameters are stored in the library of safety-critical scenarios for future generations. {\bfseries{Stage-IV}}: Collected rare-event scenarios are sampled with the proposed method to update model parameters for safer RL agent.}
    \vspace*{-4mm}
    \label{fig:system_overview}
\end{figure*}

In this work, we investigate whether the rare-event assessment and validation samples of a safety-critical system can be utilized for automatic self-improvement (healing) of the RL agents. Our specific contribution is the development of a framework (see Fig. \ref{fig:system_overview}) where the RL-based AD system is continuously subjected to probabilistic Black-box verification methods to discover failure scenarios. The proposed framework incrementally adjusts the training scenarios of the RL agent to compensate for the failures in these scenarios. We select ACC as a case study for the application of our method and show that the proposed system significantly reduces the number of safety violations.

%%%%%%%%%%%%%%%%%%%%%%%%%%%%%%%%%%%%%%%%%%%%%%%%%%%%%%%%%%%%%%%%%%%%%%%%%%%%%%%%%%%%%%%%%%%%%%%%
% ----------------------------------------- BACKGROUND ----------------------------------------- 
%%%%%%%%%%%%%%%%%%%%%%%%%%%%%%%%%%%%%%%%%%%%%%%%%%%%%%%%%%%%%%%%%%%%%%%%%%%%%%%%%%%%%%%%%%%%%%%%
\section{Background}
This section contains background information on the verification methodologies, deep RL algorithms, and model-based baseline approach utilized in this study.

\subsection{Rare-Event Simulation} \label{sec:algoritms_section}
The verification and testing of a black-box system in adversarial, safety-critical scenarios are conducted in a risk-based simulation framework as shown in Fig. \ref{fig:system_overview}. Let $\mathbf{x} \sim \mathbb{P}_0$ be scenario parameters where $\mathbb{P}_0$ is a base distribution of the simulations. We define the safety function as $\digamma: \raisebox{2pt}{$\chi$} \rightarrow \mathbb{R}$ with safety metric $\tau$ and evaluate the probability of rare events.
\begin{equation}
\label{eqn:objective}
    p_\tau := \mathbb{P}_0(\digamma(\mathbf{x}) \leq \tau)
\end{equation}

\subsubsection{Grid-Search Approach}
Grid-Search (GS) is a simple parameter search approach across a bounded search space. In GS, the continuous search space is periodically partitioned into discrete grids. The interval value and dimension of the search space heavily influence the performance. However, the GS method suffers from the curse of dimensionality.

\subsubsection{Monte-Carlo Sampling}
The na\"ive Monte-Carlo (MC) estimator probability $\mathbb{P}(\phi)$ can be defined as in Eq. (\ref{eqn:naive_mc}), where $\phi$ is a sampling scenario event, $N$ is the total number of samples from the normal distribution $\mathcal{N}$.
\begin{equation}
\label{eqn:naive_mc}
    p(\phi) \approx \frac{1}{N} \sum_{i=1}^{N} \mathbb{P}(\phi | \mathbf{x_i}), \quad \mathbf{x_i} \overset{\mathrm{iid}}{\sim} \mathbb{P}(\mathbf{x})
\end{equation}

As shown in Eq. (\ref{eqn:rare_mc}), the conditional na\"ive MC method can be applied to calculate the rare-event simulation probability $\hat{p}_\tau$, where $\tau$ is a constant TTC threshold that reflects an ACC safety/comfort metric \cite{stochastic}. The continuous objective function $\digamma: \raisebox{2pt}{$\chi$} \rightarrow \mathbb{R}$ reflects the safety of the scenarios, where low values of $\digamma(\mathbf{x})$ correlate to a higher likelihood of a collision.
\begin{equation}
\label{eqn:rare_mc}
    \hat{p}_\tau \approx \frac{1}{N} \sum_{i=1}^{N} \mathds{1}\{\digamma(\mathbf{x_i}) \le \tau\}, \quad \mathbf{x_i} \in \raisebox{2pt}{$\chi$}
\end{equation}

By using a na\"ive MC method, all rare events can be found when $N \rightarrow \infty$. However, as the number $N$ and the search space dimension increase, both the GS and MC simulation techniques become inadequate despite their simplicity.

\subsubsection{Bayesian Optimization} \label{sec:bayesian}
Adversarial AD scenarios are effectively found via Bayesian Optimization (BO) \cite{generating_adversarial}. An acquisition function $\mathbb{A}(\phi)$ in Eq. (\ref{eqn:acquisition}) is maximized at each iteration while taking the mean $\mu$ and variance $\sigma^2$ of the most recent BO predictions into account using Gaussian Process (GP) \cite{gaussian}. At each simulation rollout, scenario prediction is sampled as $y = \digamma(\phi) + \mathcal{N}$ and the optimality is discovered by minimizing an objective function as $\phi^* = \underset{\phi}{\mathrm{argmin}}[\digamma(\phi)]$.
\begin{equation}
\label{eqn:acquisition}
    \mathbb{A}(\phi) = \mathbb{E}_{\mathbb{P}(y | \phi)}[\textit{I}(\phi)], \quad \textit{I}(\phi) = \mathrm{max}(0, \digamma^* - y)
\end{equation}
where $\textit{I}(\phi)$ is an improvement function and $\digamma^*$ is a minimum value of an objective function given $\phi$ scenario sample \cite{bayesian}.

\subsubsection{Cross-Entropy Search} \label{sec:cross_entropy}
A Cross-Entropy (CE) method is one of the widely used parametric adaptive importance-sampling (AIS) strategies \cite{cross_entropy}.
\begin{equation}
\label{eqn:cross_entropy}
    \hat{p}_\tau = \frac{1}{N} \sum_{i=1}^{N} \frac{p_0(\mathbf{x_i})}{p^*(\mathbf{x_i})} \mathds{1}\{\digamma(\mathbf{x_i}) \le \tau\}, \quad \mathbf{x_i} \in \raisebox{2pt}{$\chi$}
\end{equation}
where $p_0 = p_{\tau} \cdot p^*$ is a probability density function (pdf) of a base distribution $\mathbb{P}_0$ under random variable $\raisebox{2pt}{$\chi$}$. Despite being an exact solution, Eq. (\ref{eqn:cross_entropy}) is impossible to compute since we are not sure of the definite result of likelihood ratio $p_{\tau}$. By repeatedly minimizing Kullback-Leibler projection of $p^*$ onto the $\theta$ parameterized probability distribution $p_{\theta}$ as $\theta^*$ $\in$ $\underset{\theta}{\mathrm{argmin}}[D_{KL}(p^* || p_{\theta})]$, proximal value of the probability in Eq. (\ref{eqn:cross_entropy}) is computed \cite{scalable}.

\subsubsection{Adaptive Multilevel Splitting} \label{sec:ams}
The sample inefficiency of na\"ive MC is addressed by the non-parametric Adaptive Multilevel Splitting (AMS) algorithm \cite{ams}, which computes the rare-event probability $p_{\tau}$ by dividing the safety threshold into $J$ intermediate levels, with $\infty =: \tau_0 > \tau_1 > \dots > \tau_{J-1} > \tau_J:= \tau$. The probability of each level is calculated as shown in Eq. (\ref{eqn:ams}).
\begin{equation}
\label{eqn:ams}
    \mathbb{P}_0(\digamma(\mathbf{x}) < \tau) = \prod_{j=1}^{J} \mathbb{P}(\digamma(\mathbf{x}) < \tau_j | \digamma(\mathbf{x}) < \tau_{j-1})
\end{equation}

As proposed in \cite{efficient}, each intermediate level is determined by specifying a constant fraction $\delta \in (0, 1)$ that is dropped at each intermediate iteration phase $j$.

\subsection{Deep Reinforcement Learning}
The main objective of RL under parameter $\theta$ is to maximize the future cumulative reward over trajectories $\omega = \{ s_0, a_0, r_0, ..., s_T, a_T, r_T \}$ with policy $\pi_{\theta}(a_t | s_{t})$, where action $a_t \in A$, state $s_t \in S$, and $\mathbf{R}_t$ is the reward at time $t$. The objective function is given by $J(\theta)$.
\begin{equation}
\label{eqn:rl_objective}
    J(\theta) = \mathbb{E}_{\omega \sim \pi} \left[ \sum_{t=0}^{T} \gamma^t \cdot \mathbf{R}_t \right]
\end{equation}
An advantage estimator function $\hat{A}_{t}$ is defined as in Eq. (\ref{eqn:advantage}).
\begin{equation}
\label{eqn:advantage}
    \hat{A}_{t}^{\pi_k} \approx \sum_{i = 0}^{T - 1} \big[ \gamma^i \cdot \mathbf{R}_{t + i} \big] + \gamma^n \cdot V(s_{t + T}) - V(s_t)
\end{equation}
where $V(s_t) = \mathbb{E}[G_t | s_t]$ is a value function, $G_t$ is a discounted sum of rewards and $\gamma \in [0, 1)$ is a constant discount factor that controls the weight given to future rewards.

In this work, we evaluated our system by applying the continuous action policy-gradient-based algorithm, on-policy Proximal Policy Optimization (PPO) \cite{ppo}. In an on-policy RL, the policy for choosing actions and the policy for updating the target neural network model are the same, while they are different in an off-policy RL. For our case study, we chose to use PPO exclusively as an on-policy RL algorithm.

\subsubsection{Proximal Policy Optimization}
The original version of the clipped objective function of PPO is given in Eq. (\ref{eqn:ppo}).
\begin{equation}
\label{eqn:ppo}
    \begin{split}
        J_{PPO}(\theta_k) = & \mathbb{E}_{\omega \sim \pi_k} \bigg[ \sum_{t=0}^{T} \Big[ \mathrm{min}(r_t(\theta) \cdot \hat{A}_{t}^{\pi_k}(s_t, a_t),\\
        & \mathrm{clip}(r_t(\theta), 1 - \epsilon, 1 + \epsilon) \cdot \hat{A}_{t}^{\pi_k}(s_t, a_t)) \Big] \bigg]
    \end{split}
\end{equation}
where $r_t(\theta) = \frac{\pi_{\theta_k}(a_t | s_{t})}{\pi_{\theta_{k-1}}(a_t | s_{t})}$ represents the ratio between the current policy and the previous policy. The effect of deviation from the previous policy $\pi_{k-1}$ is controlled by $\epsilon = 0.2$ value.

\subsection{Intelligent-Driver Model} \label{sec:idm}
One of the model-based methods used in ACC scenarios is the Intelligent Driver Model (IDM) \cite{idm}. The IDM generates an instantaneous longitudinal acceleration $\alpha$ of the vehicle using a continuous model, as specified in Eq. (\ref{eqn:idm}).
\begin{equation}
\label{eqn:distance}
    d^*(v, \Delta v) = d_0 + v \cdot T_{safe} + \frac{v \cdot \Delta v}{2 \cdot \sqrt{b \cdot \alpha_{max}}}
\end{equation}
\begin{equation}
\label{eqn:idm}
    \alpha = \frac{dv}{dt} = \alpha_{max} \cdot \Bigg[ 1 - \bigg( \frac{v}{v_d} \bigg)^\delta - \bigg( \frac{d^*(v, \Delta v)}{d} \bigg)^2 \Bigg]
\end{equation}
where $\alpha_{max} = 4 m/s^2$ is the maximum acceleration, $\delta = 4$ is the acceleration exponent, $d_0 = 3.0 m$ is the minimum distance gap, $b = -2.0 m/s^2$ is the desired deceleration, $T_{safe} = 1.5 s$ is the safe headway time, $d$ is the distance to the front vehicle if it exists, and $v$ and $v_d$ are the current and desired velocities in $m/s$. The maximum deceleration of the IDM is manually limited to $a_{min} = -4 m/s^2$. Other constant parameters of the model are set to be utilized as in the default simulation environment configurations \cite{highway_env}.

%%%%%%%%%%%%%%%%%%%%%%%%%%%%%%%%%%%%%%%%%%%%%%%%%%%%%%%%%%%%%%%%%%%%%%%%%%%%%%%%%%%%%%%%%%%%%%%%
% -------------------------------------- METHODOLOGY ------------------------------------------- 
%%%%%%%%%%%%%%%%%%%%%%%%%%%%%%%%%%%%%%%%%%%%%%%%%%%%%%%%%%%%%%%%%%%%%%%%%%%%%%%%%%%%%%%%%%%%%%%%
\section{Methodology}
In this section, the introduction of the problem setting, RL model structure, evaluation metrics, and continuous self-improvement methodology, shown in Fig. \ref{fig:system_overview}, is carried out.

\subsection{Problem Setting} \label{sec:problem_setting}
\begin{figure}[htbp]
    \centering
    \vspace*{0.06mm}
    \includegraphics[width=0.485\textwidth]{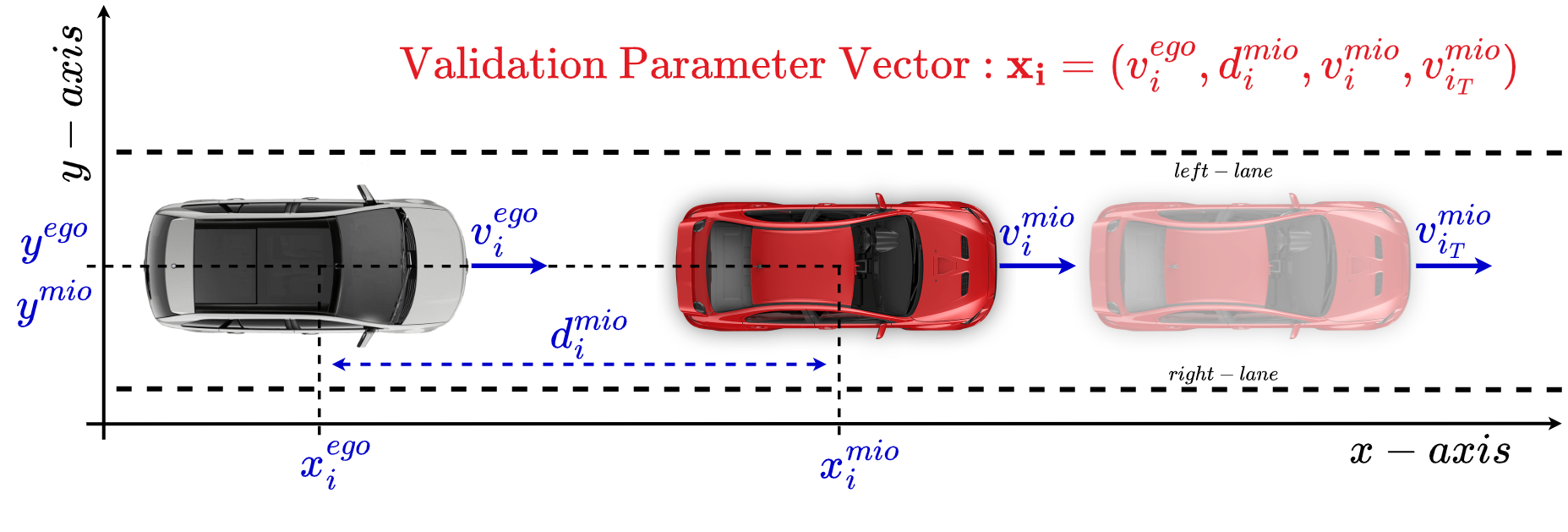}
    \vspace*{-6.3mm}
    \caption{Case study illustration of validation parameters in an ACC scenario, where $v_{i}^{ego}$ is an initial velocity of the EGO vehicle, $d_{i}^{mio}$ is an initial MIO vehicle distance relative to an EGO vehicle, $v_{i}^{mio}$ and $v_{i_T}^{mio}$ are initial and target velocities of the MIO vehicle for a scenario parameter vector $\mathbf{x_i}$.}
    \label{fig:acc_scenario}
\end{figure}

We constructed the problem of finding safety-critical scenarios in an ACC driving application using rare-event simulation techniques and proposed an effective and sample-efficient continuous self-improving framework. Initially, we visualized certain features in an ACC scenario, such as the time-gap ($\mathcal{T}^{gap}$) in Eq. (\ref{eqn:tgap}) and time-to-collision ($\tau^{ttc}$) in Eq. (\ref{eqn:ttc}) between the lead (MIO) and following (EGO) vehicles, which are illustrated in Fig. \ref{fig:acc_scenario}.
\begin{equation}
\label{eqn:tgap}
    \mathcal{T}_{t}^{gap} = \dfrac{d_{t}^{mio}}{v_{t}^{ego}} \quad \forall \text{ } t \in (0, 1, \dots, T)
\end{equation}

As acknowledged in previous studies \cite{efficient, scalable}, we chose TTC, given in Eq. (\ref{eqn:ttc}), as an objective metric to minimize in rare-event simulations. We formulated risk-based evaluation to minimize the objective function $\digamma(\mathbf{x_i})$, as low values of $\tau_{i}^{ttc}$ are rare and dangerous. $\digamma(\mathbf{x_i})$ is evaluated by simulating the black-box model, which is, in our case, an RL agent.
\begin{equation}
\label{eqn:ttc}
    \tau_{t}^{ttc} = \dfrac{d_{t}^{mio}}{v_{t}^{ego} - v_{t}^{mio}} \quad v_{t}^{ego} > v_{t}^{mio}, \text{ } t \in (0, 1, \dots, T)
\end{equation}

The objective TTC value $\tau_{i}^{ttc}$ for scenario event $\mathbf{x_i}$ is determined by simulating a black-box model until the episode is done (terminated), as defined in Eq. (\ref{eqn:metric_ttc}).
\begin{equation}
\label{eqn:metric_ttc}
    \digamma(\mathbf{x_i}) = \tau_{i}^{ttc} = \arg \min_{t} \big( \tau_{t}^{ttc} \big), \quad \forall \text{ } t \in (0, 1, \dots, T)
\end{equation}

We selected $\mathbf{x_i} = \{ (v_{i}^{ego}, d_{i}^{mio}, v_{i}^{mio}, v_{i_T}^{mio}) \in \mathbb{R}^4: \space v_{i}^{ego}, d_{i}^{mio}, v_{i}^{mio}, v_{i_T}^{mio} \in \mathbb{R}\}$ as a validation parameter vector to represent simulation parameters, where $v_{i}^{ego}$ is the initial velocity of an EGO vehicle, $v_{i}^{mio}$ is the initial velocity of the MIO vehicle, $d_{i}^{mio}$ is the initial distance to the MIO vehicle, and $v_{i_T}^{mio}$ is the desired/target velocity of the lead vehicle in a particular scenario sample $i$. The validation parameter vector elements are visually explained in Fig. \ref{fig:acc_scenario}. For our case study, we fixed the lateral movement of the vehicles at $y^{ego} = y^{mio} = 0.0 m$. Additionally, the range limits of the validation parameter vector elements are given in Table \ref{table:observations_actions}.

In this work, we used IDM to control the longitudinal motion of the MIO vehicle. The configuration parameters of the IDM are fixed throughout all experiments in this study.

\subsection{Black-Box Deep RL Model} \label{sec:deep_rl}
A black-box RL agent is trained with a continuous action policy algorithm PPO to learn longitudinal low-level control behavior in safety-critical ACC scenarios.
{
\renewcommand{\arraystretch}{1.5}
\begin{table}[ht]
\centering
\caption{The observation and action spaces of the black-box model.}
\begin{tabular}{ c || c || c }
    \hline
        \multicolumn{3}{c}{\bfseries{Observation Space} ($\mathbf{S}$)} \\
    \hline
        $v_{t}^{ego}$ & longitudinal velocity & \big[$10$m/s, $40$m/s\big] \\
    \hline
        $\dot{v_{t}}^{ego} = \alpha_{t}$ & longitudinal linear acceleration & \big[$-4$m/s$^2$, $4$m/s$^2$\big] \\
    \hline
        $\dot{\alpha_{t}}$ & EGO vehicle jerk in x-axis & \big[$-8$m/s$^3$, $8$m/s$^3$\big] \\
    \hline
        $d_{t}^{mio}$ & MIO vehicle relative distance & \big[$10$m, $120$m\big] \\
    \hline
        $v_{t}^{mio} - v_{t}^{ego}$ & MIO vehicle relative velocity & \big[$-40$m/s, $40$m/s\big] \\
    \hline
        \multicolumn{3}{c}{\bfseries{Action Space} ($\mathbf{A}$)} \\
    \hline
        $a_t$ & throttle \& brake combination & \big[$-1$, $+1$\big] \\
    \hline
\end{tabular}
\label{table:observations_actions}
\end{table}
}

\subsubsection{State Space}
The observation state $s_t$ of our RL agent at each step $t \in (0, 1, \dots, T)$, where $T$ is the episode horizon, can be represented as a generalized vector space format in Table \ref{table:observations_actions} as $\mathbf{S} = \{ s_t \in \mathbb{R}^5: v_{t}^{ego}, \alpha_{t}, \dot{\alpha_{t}}, d_{t}^{mio}, v_{t}^{mio} - v_{t}^{ego} \in \mathbb{R}\}$. Normalization is applied before feeding an observation state to a simple dense $2$-layer hidden multi-layer perceptron (MLP) model with $2 \times [64, \text{ReLU}]$ number of neurons for both the actor and critic networks.

\subsubsection{Action Space}
A control action $a_t$ of the black-box model is denoted as a combination of throttle ($a_t > 0$) and brake ($a_t < 0$) pedal openings. The output of the actor (policy) network with $tanh$ activation function is directly applied to the simulator at each episodic step $t$. We restricted our action space to a continuous domain of longitudinal control without any steering motion as our case study only involves ACC scenarios on a straight-road highway.

\subsubsection{Reward Function}
The custom reward function in Eq. (\ref{eqn:reward}) is designed to express the vehicle following scenario with a safety metric ($\mathcal{T}^{gap}$). A positive reward is given when the black-box agent follows the lead vehicle (MIO) with $0.8 s \leq \mathcal{T}^{gap} \leq 2.0 s$. However, when the agent fails to drive safely and crashes with the front vehicle, a significant penalty (negative reward) is given, and the episode is terminated. The reward function $\mathbf{R}_t(s_t, a_t)$ for RL training is as follows:
\begin{equation}
\label{eqn:reward_tgap}
    \mathbf{R}_{t}^{gap} =
        \begin{cases}
            - \dfrac{1}{\mathcal{T}_{t}^{gap}}& \quad \text{if \space} \mathcal{T}_{t}^{gap} \in \big( 0.0, \text{ } 0.8 \big)  \\
            + \mathcal{T}_{t}^{gap}           & \quad \text{if \space} \mathcal{T}_{t}^{gap} \in \big[ 0.8, \text{ } 2.0 \big]  \\
            - \mathcal{T}_{t}^{gap}           & \quad \text{if \space} \mathcal{T}_{t}^{gap} > 2.0
    \end{cases}
\end{equation}
\begin{equation}
\label{eqn:reward}
    \mathbf{R}_t(s_t, a_t) =
    \begin{cases}
        -10.0 & \text{collision} \\
        \mathbf{R}_{t}^{gap} + c_{j} \cdot \dot{\alpha}_t + \dfrac{c_{sp} \cdot v_{t}}{v_{max}} & \text{otherwise}
    \end{cases}
\end{equation}
where $c_j = -0.5$ is a penalty coefficient for the vehicle jerk $\dot{\alpha}_t$ and $c_{sp} = 5.0$ is a reward coefficient for the velocity $v_t$ at the episodic time-step $t$. These weight values are determined through experiments and account for the safety and comfort emphasis in the ACC scenarios. When the black-box agent experiences a jerk, a negative reward is provided since it closely resembles a lack of driving comfort.

\subsection{Self-Improving Method} \label{sec:self_improving_method}
\begin{figure}[htbp]
    \centering
    \vspace*{-5.0mm}
    \includegraphics[width=0.448\textwidth]{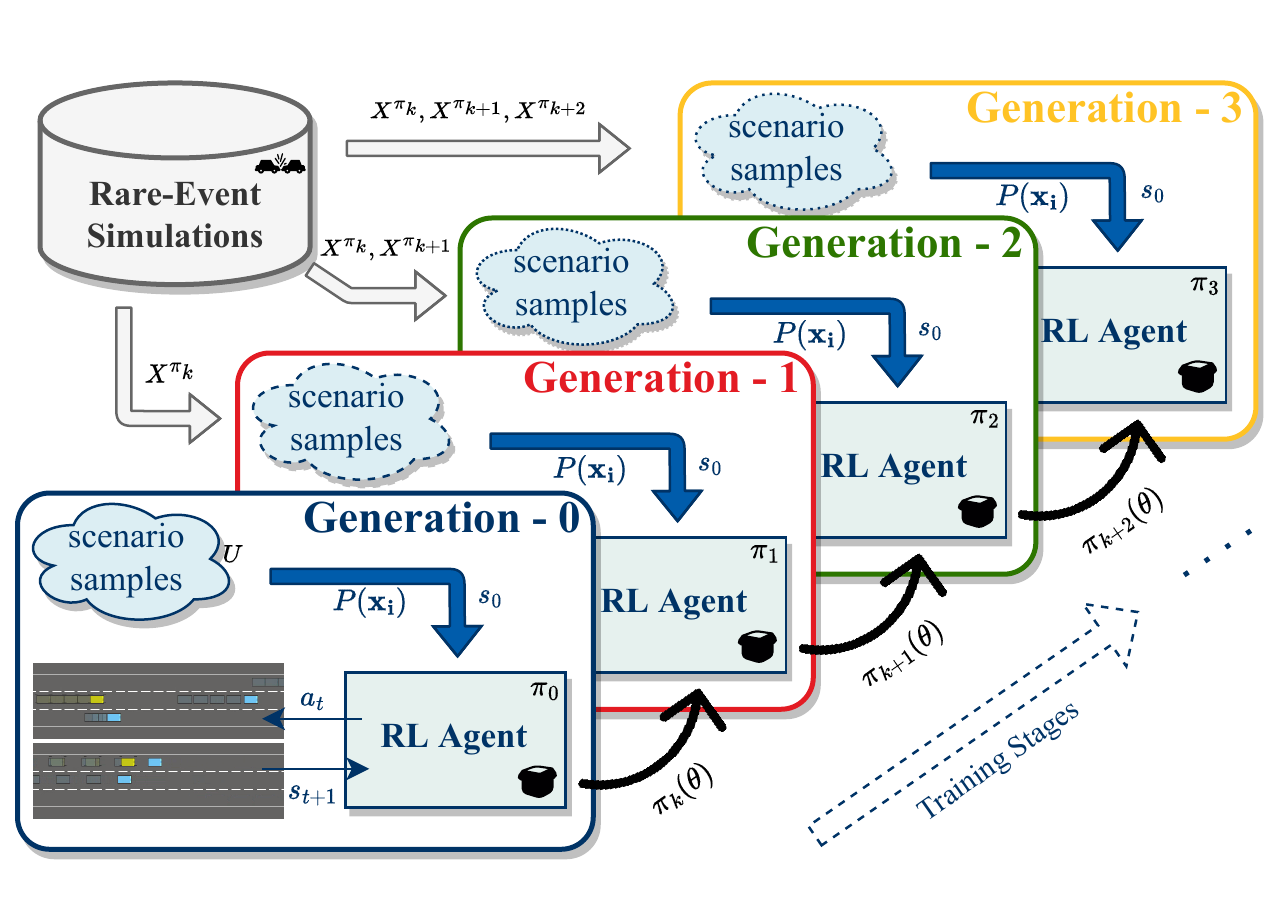}
    \vspace*{-4.5mm}
    \caption{{\bfseries{Proposed Improvement Stages}}. The safety performance of the black-box model is increased with self-improvement during RL training.}
    \vspace*{-1.2mm}
    \label{fig:self_healing}
\end{figure}

We proposed a continuous learning pipeline in which a black-box model is periodically tested with rare-event simulations and improved by effectively leveraging safety-critical scenarios during training stages. All verification samples are stored in a library to be sampled during training of the black-box agent policy. The probability of sampling from the library of rare-event simulations is given in Eq. (\ref{eqn:self_healing}) as:
\begin{equation}
\label{eqn:self_healing}
    P(X^{\pi_k} | \pi_k(\theta)) = \frac{1}{2} \cdot \dfrac{k+1}{\sum_{g=0}^{G+1}\big[ g \big]}, \quad k \in (0, 1, \dots, G)
\end{equation}
where $G$ is the total number of generations while training a black-box model. $X^{\pi_k} = \{ \mathbf{x_i} : \forall \text{ } i \in (1, 2, \dots, N)\}$ is a set of validation scenario events of the policy $\pi_k$, where $k$ is a generation number and $N$ is a total number of rare-event simulations. As illustrated in Fig. \ref{fig:self_healing}, at each generation, the trained model parameters $\theta$ are transferred to the next generation policy to distill learned knowledge and gained experience from previous generations so the continuous transfer learning is carried out. One of the rare-event simulation methods, detailed in Section \ref{sec:algoritms_section}, is applied for black-box model verification after each generation stage separately. A set of scenario simulation parameters $X^{\pi_k}$ of the verified/tested policy $\pi_k$ (black-box model) is added to the library of rare-event simulations (safety-critical scenarios) as depicted at Stage-III in Fig. \ref{fig:system_overview}. After computing the probability in Eq. (\ref{eqn:self_healing}), an initial state observation $s_0$ is determined by selecting scenario event parameter $\mathbf{x_i} \in X^{\pi_k}$ from the uniform discrete random distribution $U\{\mathbf{x_0}, \mathbf{x_N}\}$.

In the Generation-0 stage of black-box policy training, as shown in Fig. \ref{fig:self_healing}, the scenario initial condition for each training episode is constructed from an initial state observation $s_0$ that is sampled from the uniform continuous random distribution within the state space ranges provided in Table \ref{table:observations_actions}. However, in subsequent generations of black-box model training, scenarios are sampled from a combination of both a uniform distribution with a probability of $P_{U}=1/2$ and the library of rare-event simulations as in Eq. (\ref{eqn:self_healing}). The likelihood of sampling a safety-critical scenario $P(X^{\pi_k})$ from generation $k$ is increased compared to previous generations, as explained in Eq. (\ref{eqn:self_healing}), as the aim is to enable the training agent to learn from its mistakes and safety-critical failures.
\begin{figure*}[t!]
    \centering
    \includegraphics[width=0.968\textwidth]{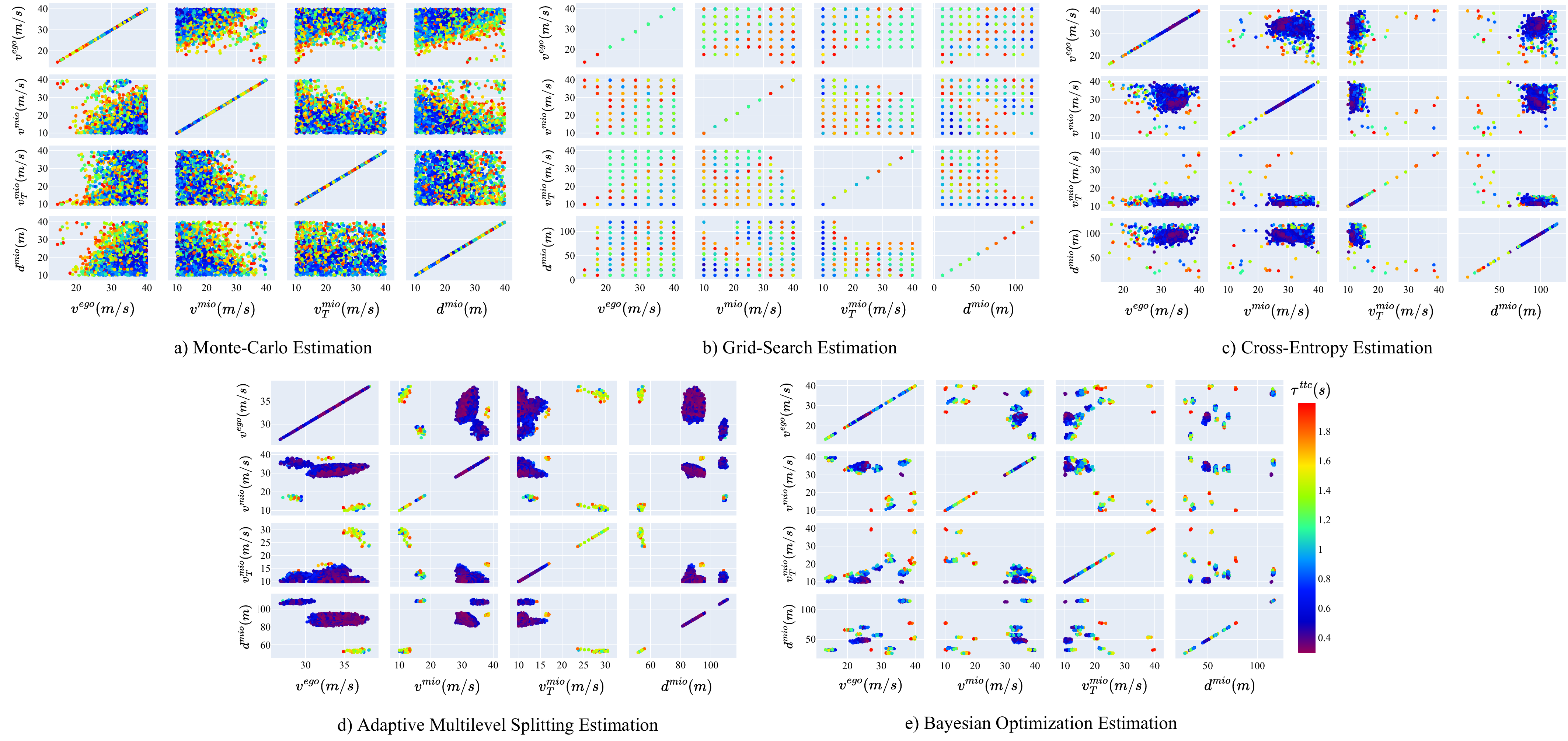}
    \caption{{\bfseries{Generating Safety-Critical Scenarios}}. The results of the rare-event scenario estimation algorithms for $\tau^{ttc} \in \big[ 0.0 s, 2.0 s \big ]$ are visualized in matrices. Na\"ive MC (a) and GS (b) sampling methods are inefficient for estimating the rare-event probability $p_{\tau}$. A parametric CE (c) method converges to one local minimum, while non-parametric AMS (d) and BO (e) approaches estimate importance sampling distribution with multiple local minimums.}
    \vspace*{-3mm}
    \label{fig:validation_scatters}
\end{figure*}

%%%%%%%%%%%%%%%%%%%%%%%%%%%%%%%%%%%%%%%%%%%%%%%%%%%%%%%%%%%%%%%%%%%%%%%%%%%%%%%%%%%%%%%%%%%%%%%%
% --------------------------------- EXPERIMENTS & RESULTS--------------------------------------- 
%%%%%%%%%%%%%%%%%%%%%%%%%%%%%%%%%%%%%%%%%%%%%%%%%%%%%%%%%%%%%%%%%%%%%%%%%%%%%%%%%%%%%%%%%%%%%%%%
\section{Experiments \& Results}
In this section, the performance outcomes of several verification techniques, detailed in Section \ref{sec:algoritms_section}, are discussed along with the rare-event probabilities $p_\tau$ of the algorithms. The effectiveness of each rare-event algorithm on the safety performance of the self-improving approach, for building safe and reliable black-box systems, is analyzed in detail.

\subsection{Simulation Setup}
The training and validation simulations are performed in a Highway-Env Gym environment \cite{highway_env} since the simulator is robust, parallelizable, and sample efficient for conducting multiple-parameter traffic scenarios. As clarified in Section \ref{sec:problem_setting} and in Fig. \ref{fig:acc_scenario}, the simulator is initialized with scenario parameters $\mathbf{x_i}$ at each verification iteration $i \in (1, 2, \dots, N)$ during application of the black-box validation algorithm. Every simulation is carried out at a $10hz$ frequency with $T=250$ time-steps episode length, yielding $25s$ of scenario events. We limited the number of vehicles in an environment to two, as the ACC scenarios with only longitudinal control are studied in this work.

\subsection{Training RL Agent}
The continuous action PPO policies are trained using the Ray-RLlib library \cite{rllib}, with default configuration hyperparameters ($\gamma=0.95, \alpha_{lr}=0.0003, \epsilon=0.2, \beta_{KL}=0.01, \lambda=1.0$). The RL agent operates as a black-box model, subject to continual verification and testing using rare-event estimation algorithms, as shown in Fig. \ref{fig:system_overview}. At each continual generation, the trained parameters of the black-box model are loaded and the training process continues. The training and validation environment of the black-box model is depicted in Stage-IV of Fig. \ref{fig:system_overview}, and one generation is defined as a single continual loop of self-improvement, as illustrated in Fig. \ref{fig:self_healing}.

In Section \ref{sec:deep_rl}, we thoroughly described the observation space $\mathbf{S}$ and the action space $\mathbf{A}$ of a black-box RL model, and clarify their respective ranges in Table \ref{table:observations_actions}. During each generation cycle shown in Fig. \ref{fig:system_overview}, the RL agent undergoes transfer learning for $300K$ training steps. Initially, the black-box model parameters are loaded randomly, and the policy is trained by sampling from a continuous uniform distribution within the observation space ranges. After each generation, the previously trained model parameters and weights are loaded to continue optimizing the black-box agent on the upgraded rare-event scenario library, as visualized in Fig. \ref{fig:self_healing}. At the end of each episode, the environment is reset to the initial configurations of the ACC events, which are sampled from both the uniform distribution and the library of safety-critical scenarios (as described in Section \ref{sec:self_improving_method}).

\subsection{Agent Verification}
Several rare-event estimation algorithms, including the BO method, CE search, and AMS method, are applied to evaluate the black-box model continuously. Additionally, for benchmarking and comparison purposes, GS and na\"ive MC sampling techniques are tested. The sample-efficient performance of generating safety-critical scenarios that cause a black-box system to fail is depicted in Fig. \ref{fig:validation_scatters}. The objective function chosen to minimize is TTC, as ACC scenarios with lower values of TTC have a higher probability of causing vehicle accidents. A fixed number of rare-event simulation samples, $N=8192$, is used in each generation for all estimation techniques. In each generation, as visualized in Fig. \ref{fig:self_healing}, $N$ estimation samples are added to the library of rare-event simulations. Each estimation sample is formulated as a vector $\mathbf{x_i}$ and stored in the library separately for each black-box verification algorithm. A detailed explanation of the validation vector $\mathbf{x_i}$ is provided in Section \ref{sec:problem_setting}.

\subsection{Self-Improvement Results} \label{sec:self_healing_results}
\begin{figure}[htbp]
    \centering
    \vspace*{-1.0mm}
    \includegraphics[width=0.485\textwidth]{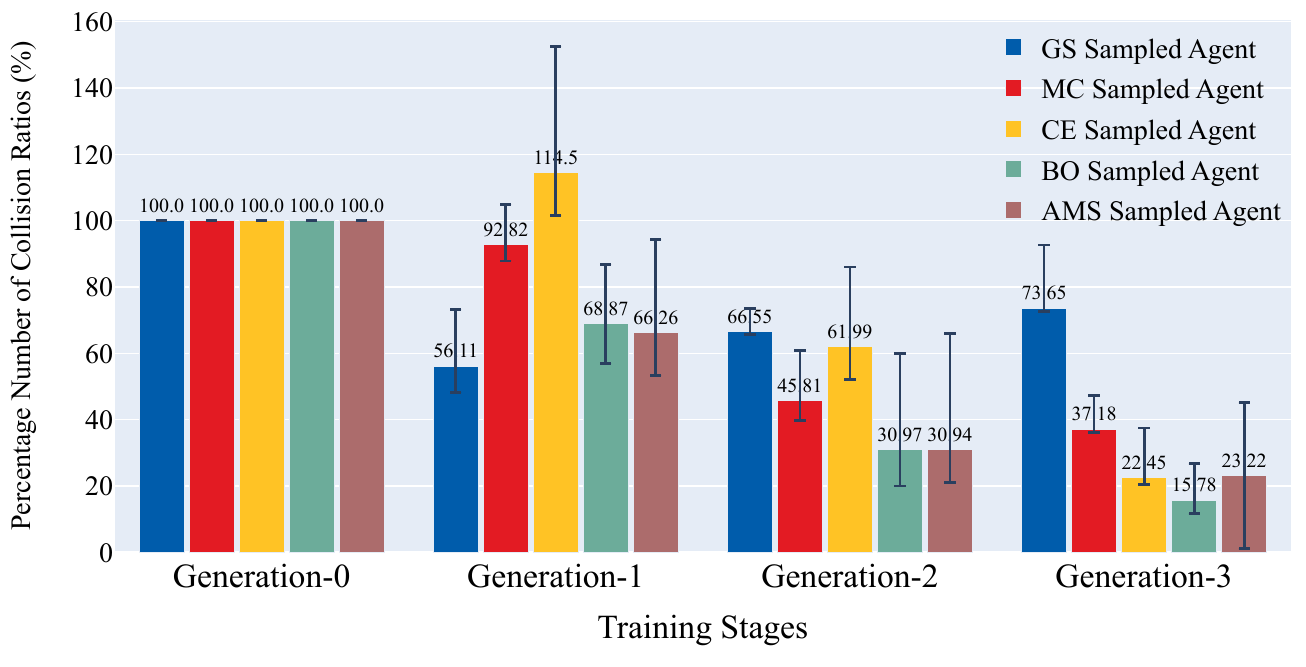}
    \vspace*{-7.0mm}
    \caption{{\bfseries{Percentage Ratios of Vehicle Collisions}}. The self-improving black-box models are evaluated at each training generation stage under a uniform state distribution. The max-min performance with the average collision ratios, normalized w.r.t. the collisions in Generation-0, is illustrated for each RL agent. The evaluations are carried out with $7$ random seeds.}
    \label{fig:collision_number_bar}
    \vspace*{-1.0mm}
\end{figure}

We evaluated the proposed self-improvement approach by training four different RL agents over three generations, totaling $1.2M$ simulation time steps for each agent. The primary objective was to decrease the number of vehicle collisions and improve the safety of the black-box system. To demonstrate the effectiveness of rare-event simulation algorithms in a self-improvement approach for a safety-critical black-box system, we compared our proposed approach with GS and na\"ive MC sampling methods using black-box verification algorithms. Fig. \ref{fig:collision_number_bar} shows the normalized number of collisions with MIO vehicles in the ACC scenarios constructed from a continuous uniform distribution space. After each proceeding generation of self-improvement with rare-event simulation samples, the black-box agent experiences fewer collisions, as demonstrated in Fig. \ref{fig:collision_number_bar} and Table \ref{table:number_of_collisions}.

\begin{figure}[htbp]
    \centering
    %\vspace*{-1.0mm}
    \includegraphics[width=0.485\textwidth]{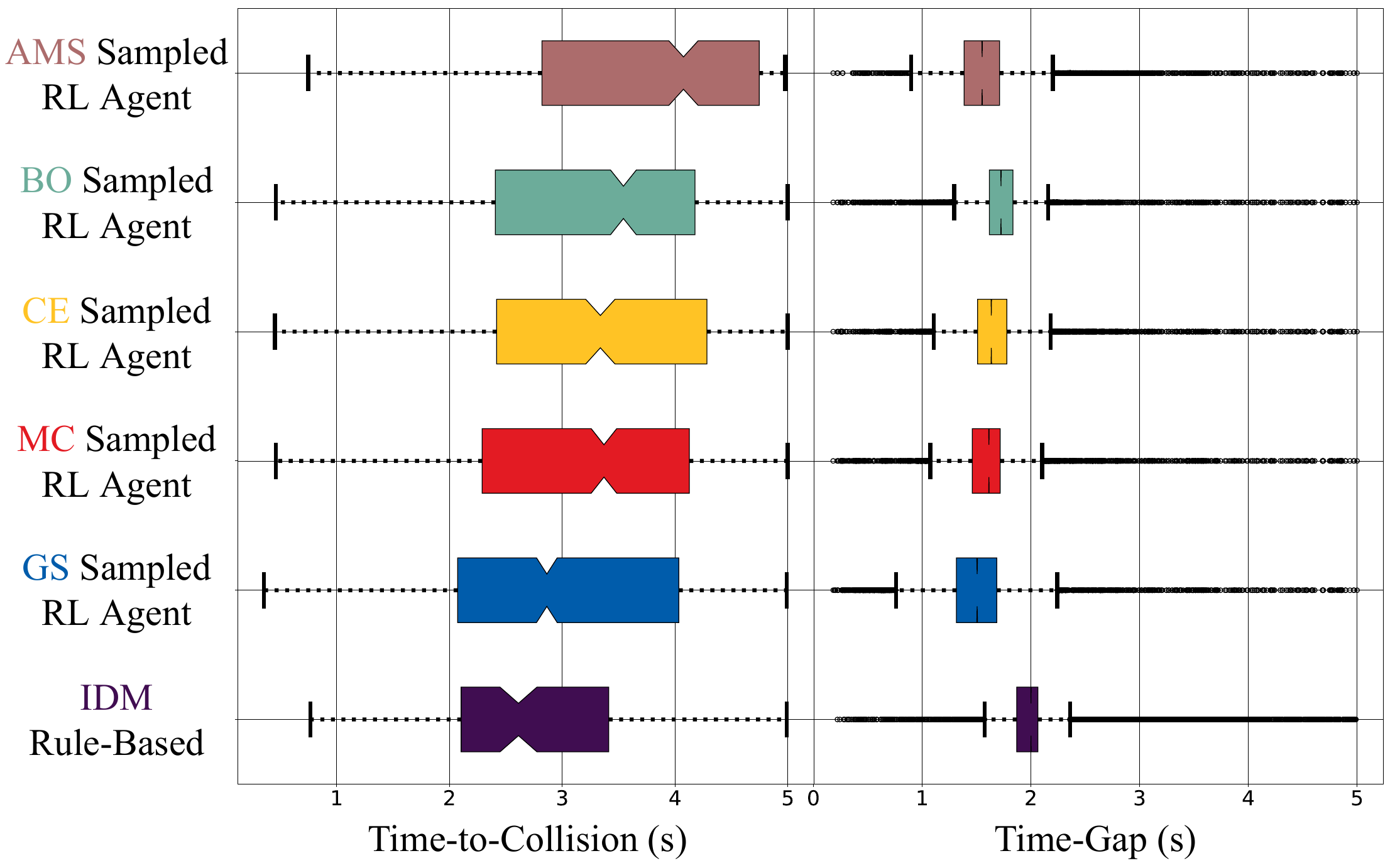}
    \vspace*{-6.0mm}
    \caption{{\bfseries{Time-Gap and Time-to-Collision Performance}}. Evaluation comparisons of the Generation-3 RL agents and the rule-based IDM in randomly selected $1000$ rare-event scenarios where $\tau^{ttc} < 5 s$.}
    \label{fig:box_plot}
    \vspace*{-1.5mm}
\end{figure}

To further investigate the safety and reliability performance of the proposed self-improvement approach, we evaluated the Generation-3 black-box RL agents on unseen safety-critical scenarios where $\tau^{ttc} < 5.0$ seconds. We chose a TTC threshold of $\tau^{ttc} < 5.0$ seconds to resemble the higher collision risk in the scenario event, as research in \cite{avoid_collisions} considers $\tau^{ttc} \in [3.5s, 5.0s]$ as a safety-critical condition. We collected $1000$ scenario events by applying CE, BO, and AMS algorithms to all Generation-3 trained RL agents and selecting $\tau^{ttc} < 5$ seconds samples randomly. We evaluated RL agents on these rare-event scenarios with $5$ different random seeds, and the collision results with standard deviations are presented in Table \ref{table:number_of_collisions}. Additionally, we evaluated a rule-based IDM, described in Section \ref{sec:idm}, on these $1000$ risky scenarios and visualized the time-gap and TTC performance comparisons in Fig. \ref{fig:box_plot}. The IDM vehicle did not crash in any of these ACC scenarios. However, the observed mean TTC values were lower than those of the trained RL agents.
{
\renewcommand{\arraystretch}{1.5}
\vspace*{-1.5mm}
\begin{table}[ht]
\centering
\caption{The Number of Collisions in Safety-Critical Scenarios.}
\vspace*{-1.0mm}
    \begin{tabular}{c || c || c || c || c}
    \cline{1-5}
        \multicolumn{5}{c}{Self-Improved RL Agents in $1000$ Samples : $\tau^{ttc} < 5 s$} \\
    \hline \hline
        {\bfseries{GS}} & {\bfseries{MC}} & {\bfseries{CE}} & {\bfseries{BO}} & {\bfseries{AMS}} \\
    \hline
        328$\pm$21 & 182$\pm$15 & 105$\pm$25 & 137$\pm$17 & 27$\pm$12 \\
    \hline
    \end{tabular}
\label{table:number_of_collisions}
\end{table}
}

\begin{figure}[htbp]
    \centering
    \includegraphics[width=0.4865\textwidth]{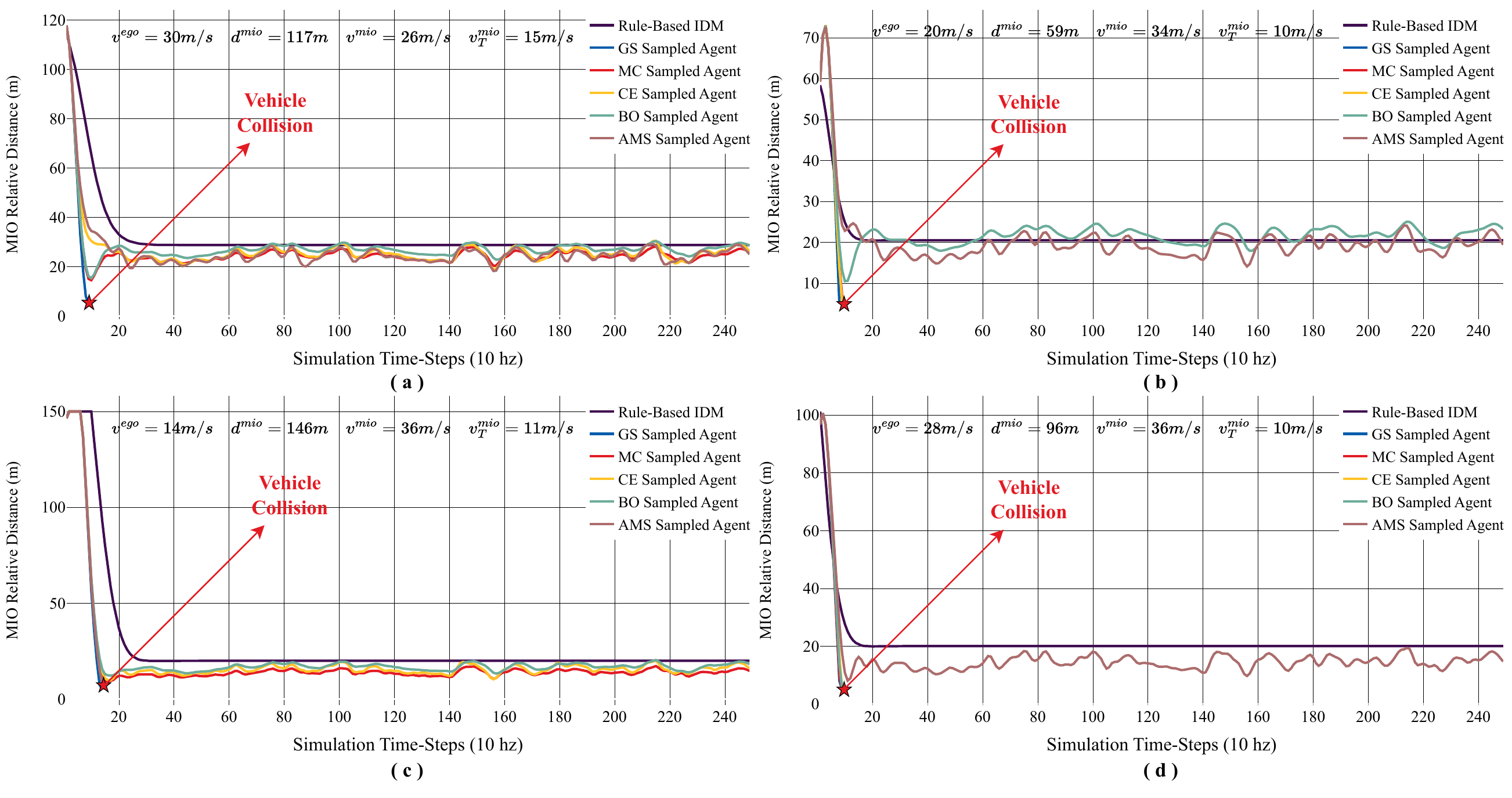}
    \vspace*{-6.0mm}
    \caption{{\bfseries{Comparison of Agent Collision Scenarios}}. Lead vehicle relative positions with respect to the EGO vehicle are compared at each time step. Equations in each plot indicate the sampled scenario parameter vector $\mathbf{x_i}$ elements (rounded here).}
    \label{fig:trajectory_plots}
    \vspace*{-4.0mm}
\end{figure}

The trajectories in Fig. \ref{fig:trajectory_plots} depict the MIO vehicle following performance of Generation-3 agents on some failure events of Table \ref{table:number_of_collisions}.

%%%%%%%%%%%%%%%%%%%%%%%%%%%%%%%%%%%%%%%%%%%%%%%%%%%%%%%%%%%%%%%%%%%%%%%%%%%%%%%%%%%%%%%%%%%%%%%%
% ---------------------------------- CONCLUSION & FUTURE WORK ---------------------------------- 
%%%%%%%%%%%%%%%%%%%%%%%%%%%%%%%%%%%%%%%%%%%%%%%%%%%%%%%%%%%%%%%%%%%%%%%%%%%%%%%%%%%%%%%%%%%%%%%%
\section{Conclusion \& Future Work}
In this work, we have proposed a novel self-improving framework using black-box verification algorithms for enhancing the safety performance of reinforcement learning-based autonomous driving agents. We have demonstrated that the weaknesses of the exploration nature of RL agents could be inspected by leveraging rare-event simulations. With the proposed methodology, safety-critical scenarios of the black-box system are detected in the early stages of the model training. Simulation results show that with our proposed iterative self-improvement approach, the RL-based black-box AD system significantly reduces failures and vehicle collisions.

The future work will expand to a higher-dimensional scenario vector space, including multiple vehicle parameters. Additionally, steering control will be added to the action space to enable more generalized driving maneuvers.

%%%%%%%%%%%%%%%%%%%%%%%%%%%%%%%%%%%%%%%%%%%%%%%%%%%%%%%%%%%%%%%%%%%%%%%%%%%%%%%%%%%%%%%%%%%%%%%%
% -------------------------------------- ACKNOWLEDGMENT ---------------------------------------- 
%%%%%%%%%%%%%%%%%%%%%%%%%%%%%%%%%%%%%%%%%%%%%%%%%%%%%%%%%%%%%%%%%%%%%%%%%%%%%%%%%%%%%%%%%%%%%%%%
\section*{Acknowledgement}
This work is supported by Istanbul Technical University BAP Grant NO: MOA-2019-42321. We gratefully thank Eatron Technologies for their technical support.

%%%%%%%%%%%%%%%%%%%%%%%%%%%%%%%%%%%%%%%%%%%%%%%%%%%%%%%%%%%%%%%%%%%%%%%%%%%%%%%%%%%%%%%%%%%%%%%%
% ---------------------------------------- REFERENCES ------------------------------------------ 
%%%%%%%%%%%%%%%%%%%%%%%%%%%%%%%%%%%%%%%%%%%%%%%%%%%%%%%%%%%%%%%%%%%%%%%%%%%%%%%%%%%%%%%%%%%%%%%%
\bibliography{references}

\addtolength{\textheight}{-12cm}  % This command serves to balance the column lengths
                                  % the text height of the last page by a suitable amount.
                                  % sure that you do not shorten the text height too much.
%%%%%%%%%%%%%%%%%%%%%%%%%%%%%%%%%%%%%%%%%%%%%%%%%%%%%%%%%%%%%%%%%%%%%%%%%%%%%%%%%%%%%%%%%%%%%%%%
\end{document}